# Identification of Parallel Passages Across a Large Hebrew/Aramaic Corpus

Avi Shmidman[1], Moshe Koppel[2], Ely Porat[3]

[1] Department of Hebrew Literature, Bar-Ilan University, Israel, and Dicta: The Israel Center for Text Analysis

[2] Department of Computer Science, Bar-Ilan University, Israel, and Dicta: The Israel Center for Text Analysis

[3] Department of Computer Science, Bar-Ilan University, Israel, and Dicta: The Israel Center for Text Analysis

**Abstract**

We propose a method for efficiently finding all parallel passages in a large corpus, even if the passages are not quite identical due to rephrasing and orthographic variation. The key ideas are the representation of each word in the corpus by its two most infrequent letters, finding matched pairs of strings of four or five words that differ by at most one word and then identifying clusters of such matched pairs. Using this method, over 4600 parallel pairs of passages were identified in the Babylonian Talmud, a Hebrew-Aramaic corpus of over 1.8 million words, in just over 11 seconds. Empirical comparisons on sample data indicate that the coverage obtained by our method is essentially the same as that obtained using slow exhaustive methods.

## INTRODUCTION

Ancient text corpora in classical languages such as Greek, Latin, Hebrew and Aramaic typically include numerous examples of text reuse, including repetitions of long passages of 20 words or more. Identifying such passages is important because it allows scholars to trace the development of ideas and concepts through time and across geographical ranges. Additionally, even within a given time period and geographical location, the identification of multiple parallel sources for any given idea provides a platform for scholarly inquiry. Identifying all examples of text reuse within a large such corpus is challenging for several reasons, including the large number of comparisons that must be done and the fact that matches tend to be only approximate.

In this paper, we specifically consider text reuse within Hebrew and Aramaic corpora. Instances of text reuse abound in classical Jewish texts, comprised of millions of words drawn from both Hebrew and Aramaic. Substantial passages of length 20 words or more recur often within this corpus in multiple texts, or within a single text in multiple contexts.

It should be emphasized that very few of these parallel passages are identical letter-for-letter parallels. First of all, any given word may be replaced by a synonymous word or phrase, and many of the passages contain interpolations of additional material. Additionally, there is much variation in the use of grammatical particles and prepositions. Finally, on the level of the individual word, orthographic variations abound. Classical Jewish texts are not at all consistent with regard to the use of *matres lectionis*; for example, any given instance of the vowel "o" may or may not be represented by the letter vuv (ו), and any given instance of the vowel "i" may or may not be represented by the letter yod (י). Additional spelling variations include alternations between final mem (ם) and final nun (ן), alternations between the letter aleph (א) and the letter yod (י), and many others. Thus, even if an entire passage were to recur word for word, a standard search algorithm would generally still miss it, due to the alterations in orthography.

One example, upon which we will focus throughout the current paper, is the Babylonian Talmud. The Babylonian Talmud contains roughly 1.8 million words, divided into 37 tractates, and includes extensive discussions of all aspects of Jewish law [Skolnick, 2007, 19:469-481].[1] The editing of the text is dated to the second half of the first millennium C.E. Thousands of passages recur within the Talmud in two or more tractates, or in multiple places within a single tractate. The identification of these parallel passages is of much significance for the study of Jewish law, because a full understanding of any given concept must take into account all of the Talmudic sections in which the concept is discussed. Indeed, before the advent of computerized databases, considerable effort was expended in compiling cross-indexes to this material by hand. The most prominent example of such is the work of Rabbi Isaiah Berlin (1725-1799), who, building upon previous work, compiled a comprehensive set of parallel Talmudic passages entitled *Masoret haShas* [Skolnick, 2007 3:456-457]. This work was printed as a cross-reference index in the margin of the classic Vilna edition of the Babylonian Talmud, and continues to be printed in virtually all modern editions of the Talmud.

For a typical example of the types of variants found within parallel Talmudic passages, consider the following passage that occurs in Tractate Shabbat and then again in Tractate Hagigah:

| Tractate Shabbat | Tractate Hagigah |
|---|---|
| ואמר רבא | והאמר רבא |
| לא חרבה ירושלים אלא בשביל | לא חרבה ירושלים עד |
| שפסקו ממנה אנשי אמנה | שפסקו ממנה בעלי אמנה |
| שנאמר שוטטו בחוצות ירושלים וראו נא | שנאמר שוטטו בחוצות ירושלם וראו נא |
| ובקשו ברחובותיה אם תמצאו איש | ובקשו ברחובותיה אם תמצאו איש |
| עושה משפט מבקש אמונה ואסלח לה | אם יש עושה משפט מבקש אמונה ואסלח לה |

Table 1. A typical pair of parallel passages and the differences between them

Within this 28/29 word parallel, we can point to five different discrepancies:

- On the first line, the Shabbat passage starts simply ואמר, while the Hagigah passage adds the interrogative particle ה, arriving at the form והאמר.
- On the second line, the Hagigah passage ends with the preposition עד, while the Shabbat passages substitutes אלא בשביל instead.
- On the third line, we find a synonym alternation: the Shabbat passage writes אנשי, while the Hagigah passage writes בעלי, although the effective meaning is identical.

- On the fourth line, while quoting a verse from Jeremiah 5:1 the Shabbat passage writes out the word ירושלים with its full plene spelling, while the Hagigah passage adopts the defective spelling found in Scripture.
- On the sixth line, the Hagigah passage quotes the Jeremiah verse verbatim, while the Shabbat passage omits the redundant words אם יש from the verse.

A traditional approach to the problem of finding approximate matches of this nature is the use of an edit-distance calculation, such as the Levenshtein distance. The edit distance between two strings is a measure of the number of transformations necessary to mutate the first string into the second. If the edit-distance is beneath a pre-defined threshold, the pair of strings can be considered a match. However, in the situation we are dealing with here – in which we want to compare all passages against all other passages – the cost of the calculation of edit-distance is prohibitive. For instance, an all-against-all comparison of 20-word pairs within a 1.8 million word corpus (the size of the aforementioned Babylonian Talmud) would require approximately 1.6 trillion calculations. According to our measurements on an Intel Core i7 processor running at 3.47 gigahertz, the computation of the Levenshtein distance for two 20-word passages requires an average of 380 microseconds. Thus, the computation of all 1.6 trillion pairs would require nearly 20 years.[2]

A number of recent studies have discussed this challenge, suggesting various heuristics and methods to reduce the computation time [Lee, 2007; Barron-Cedeno et al, 2010; Büchler 2013; Büchler 2013; Smith et al, 2013; Klein et al, 2014; Franzini et al, 2014]. It should be noted that methods will differ depending on the specific type of text reuse problem under consideration. For instance, if the goal is to identify all stylistic integrations of earlier canonical phrases within a given text, we wouldn't be interested in 20-word passages, but rather we would be looking for significant 3-word phrases, and perhaps even 2-word phrases, with only minimal variation from the source. On the other hand, when looking for reuse of whole textual units of claims, ideas, or argumentation, we will be interested specifically in longer textual sequences, allowing for a much greater breath of variances, including interpolated phrases and synonym substitutions. Each of these tasks presents its own challenges. The present paper specifically targets the latter task. We present a new, efficient and scalable method of identifying approximate sentence-length matches within Hebrew and Aramaic texts. Our method allows for variations between the passages both on the word level (orthographic differences), as well as on the phrase level (interchanged words, interpolations). Our method computes in linear time, returning all of the parallel passages throughout a corpus of arbitrary size.

## I  DESCRIPTION OF THE ALGORITHM

### 1.1 Per-word Preprocessing
As noted above, matching words in parallel Hebrew segments often differ in their orthography. In order to allow our algorithm to focus on the core of the words, while eliding the unstable parts of the words, we perform the following preprocessing procedure:

- We calculate the frequency of the Hebrew letters throughout the input corpus.

---

[2] To be sure, there are methods of speeding up the Levenshtein calculation. For an optimization using dynamic programming and parallel processing, see [Klein et al, 2014]. Another possible optimization would be to use a bag-of-letters measure as a pre-filter, automatically rejecting all pairs that exceed a certain threshold of discrepancy when comparing their bag-of-letter counts.

- For every word, we identify the two most infrequent letters, and we represent the word via those two letters (maintaining the order in which the two letters appear within the word).

Naturally, the letters representing prefixes and *matres lectionis* are among the most common the language; thus, by eliminating all but the two most infrequent letters, we effectively eliminate most prefixes and *matres lectionis*.

As an example, here is the frequency chart for Hebrew letters in the Babylonian Talmud, in order of frequency (final letters are counted together with their corresponding regular forms):

| Letter | Frequency | Letter | Frequency |
|---|---|---|---|
| י | 870162 | ד | 246182 |
| א | 696141 | כ | 218006 |
| ו | 693571 | ע | 191694 |
| מ | 581737 | ח | 157886 |
| ה | 498030 | ק | 129819 |
| ל | 488497 | פ | 113819 |
| ר | 470119 | ס | 74607 |
| נ | 416467 | ט | 64970 |
| ב | 409466 | ז | 62967 |
| ש | 301634 | ג | 59735 |
| ת | 301153 | צ | 58779 |

Table 2. The Hebrew letters and their frequencies in the Babylonian Talmud

Consider the example of דילמא and דלמא, orthographic variants of the same word. In this case, the י, א, and מ letters are eliminated, leaving only דל. The two words are thus represented by the same two letters, in the same order, allowing the algorithm to easily identify the match. The same occurs with almost all such variant pairs. Occasionally, however, the letter distribution is such that words will not be equated as expected. For instance, in the case of דתניא and לכדתניא, the first is reduced to דת, and the second to כד. Nevertheless, these cases are a minority, and they will be handled by the next step, which allows for word alternations and interpolations.

Of course, many different Talmudic words will be reduced to the same two letters, and the reader may well wonder whether this might lead to too many false positives. Nevertheless, because our goal is to find sequences of matching two-letter codes, rather than just individual words, the false positives will naturally fall by the wayside. This is demonstrated in the following chart, which details the results of examining the entire text of the Babylonian Talmud against itself based upon two-letter reductions. Every match was run against a Levenshtein filter to evaluate its validity; if the Levenshtein distance totaled greater than 20 percent of the length of the first string, then the match was considered a false positive.

| Passage Length | Valid Matches | False Positives |
|---|---|---|
| 4 | 2432511 | 1190337 |
| 5 | 733991 | 74943 |
| 6 | 294393 | 6895 |
| 7 | 189394 | 1474 |

                http://jdmdh.episciences.org

| 8 | 137145 | 377 |
| 9 | 97925 | 93 |
| 10 | 82851 | 35 |
| 11 | 72385 | 15 |
| 12 | 63750 | 12 |

Table 3. An Evaluation of the Accuracy of a Two-letter Reduction

As demonstrated here, once the passage length is sufficiently high, the two-letter-reduction method provides a very high degree of accuracy.

## 1.2 N-grams and Skip-grams

The basic unit of comparison within our algorithm is an n-gram of length 4; that is, sequences of four words. Indeed, many previous studies utilize n-grams for the identification of repeated text. However, in addition to allowing for orthographic variations within word sequences, we also need to allow for word alternations and interpolations, in which the parallel word sequences contain a word in the middle which is completely different, or which exists in one sequence but not in the other. Therefore, instead of limiting our algorithm to traditional n-grams, we use a set of noncontiguous n-grams, each of which omits one word from the text. These non-contiguous n-grams are termed "skip-grams". The effectiveness of skip-grams in identifying parallel textual segments has been demonstrated in recent studies; see for instance [Guthrie et al, 2006]. Identifying skip-grams in a text is special case of the k-mismatch problem in approximate pattern matching for which surprisingly efficient algorithms have been found [Porat and Porat, 2009].

For every 5-word sequence in the text, we extract all combinations of length 4, allowing for any of the last four words to be omitted (we don't need to include the case in which the first word is omitted, because that case is covered within the skip-gram set of the subsequent starting position). Thus, for every starting position 'x' within the corpus, we extract 4 skip-grams:

- x, x+2, x+3, x+4     [Skipping x+1]
- x, x+1, x+3, x+4     [Skipping x+2]
- x, x+1, x+2, x+4     [Skipping x+3]
- x, x+1, x+2, x+3     [Skipping x+4; effectively a contiguous 4-gram]

To take a real example, in Tractate Avodah Zarah we find the following sequence: וידע דעת עליון אפשר דעת בהמתו. A parallel passage in Tractate Berakhot states: וידע דעת עליון השתא דעת בהמתו. The respective first words of the passages – וידע and וידע – are orthographic variants equalized by the two-letter reduction. However, the fourth word of the passage is completely different (even though the import of the two words is effectively the same): השתא instead of אפשר. Nevertheless, the two passages will match up due to the third skip-gram, which elides the fourth word:

| | Original Text | Skip-grams (omitting word 3) |
|---|---|---|
| **Tractate Avodah Zarah** | וידע דעת עליון אפשר דעת | דע דע עב דע |
| **Tractate Berakhot** | וידע דעת עליון השתא דעת | דע דע עב דע |

Table 4. Skip-grams: A Practical Example

The differing words can also be in disparate places within the two passages. Similarly, this method will successfully match any pair of segments, one of length 4 and one of length 5, in which an extra word appears in the latter but not in the former.

## 1.3 Indexing the Skip-grams

As a general rule, n-gram-based algorithms would, at this point, proceed to compute a Rabin-Karp numerical hash of each n-gram [Karp and Rabin, 1987], in order to index them for quick retrieval. The use of such hashes can, however, slow down performance due to potential collisions. Fortunately, we can leverage our two-letter reduction method (detailed above, section 1.1) in order to efficiently index our skip-grams without any hash collisions. As noted in the previous section, every skip-gram consists of four words, each represented by two letters. There are a total of 22 letters in the Hebrew and Aramaic languages (we count final letters as regular forms), and, since each skip-gram is effectively a permutation of 8 such letters, there are a total of $22^8$ possibilities – just under 55 billion. We can thus uniquely represent each skip-gram with a single 64-bit integer. For efficiency, however, we maintain a set of 484 skip-gram indexes – corresponding with the 484 (=$22^2$) possibilities for the first word of the skip-gram; the $22^6$ possibilities for the remaining six letters can be uniquely represented via 32-bit integers. The limited number of possibilities for each part of the skip-grams also allows us to utilize precomputed tables in calculating their numeric representations. Thus, the entire set of skip-grams for our corpus – over 7 million in all – can be converted quickly and efficiently into an indexed numeric structure, allowing immediate access to all matches for any given skip-gram, without any need to check for hash collisions.

Once the indexes have been prepared, we can identify all matching skip-grams for any given position in the text via the following steps:

1. Retrieve the numeric representation of the two-letter code for the word at this position (a number between 0 and 483). This number serves as an offset into our set of indexes, delineating the index that we use in step 3 below.

2. For each of the four skip-grams at this position, retrieve the numerical representation of the remaining 6 letters.

3. Using the index identified in the first step, perform a lookup of the numerical representations of each of the four 6-letter patterns. Retrieve all of the skip-grams listed for each of the patterns (ignoring any that are in the immediate vicinity of the current word position).

This procedure provides the set of all relevant matching skip-grams.

## 1.4 Identifying clusters of skip-grams

Of course, a skip-gram match alone does not necessarily indicate a parallel passage. As we saw above, our skip-grams are only 4 words long, and, furthermore, they are based upon two-letter reductions of the words. What we really need to find, in order to validate a given match, is a cluster of matching skip-grams [Grozea et al, 2009]. For our purposes, we define a cluster as follows: a set of $i$ or more matching skip-grams with gaps of no more than $j$ words in between the skip-grams, stretching across a total of at least $k$ words from the start of the first skip-gram to the end of the last one. For this paper, we use the values $i$=3, $j$=8, $k$=20.

One important benefit of this method is that we will succeed in matching passages in which phrases up to 8 words in length are interpolated in one or the others of the passages, or in which variant strings of up to 8 words appear in both of the passages.

In order to efficiently identify clusters of skipgram matches, we generate a two-dimensional graph, wherein each skipgram match is plotted on one axis according to the starting word position of the base skipgram, and on the other axis according to the starting word position of the matching skipgram, similar to a dotplot [Basile et al, 2009; Grozea et al, 2011]. We are interested in the cases in which multiple skipgram matches cluster on a more-or-less diagonal line on the graph. To efficiently find such cases, we bin the skipgrams based upon the difference between their two coordinates. We review the bins which contain multiple skipgrams and consider whether those skipgrams can cluster together to form a match as per the criteria defined above. When a skipgram cluster does match the specific criteria, then we add the corresponding passages to our set of final results.

## II  RESULTS

Our algorithm, running in a single thread, takes just over 11 seconds to complete its analysis of the entire Babylonian Talmud, including the preprocessing, the indexing and skip-gram matching, and the identification of clusters. This is a drastic improvement over the many thousands of hours that would have been required to arrive at the same results via edit-distance calculations.

Overall, our algorithm identified a total of 4602 pairs of parallel passages in the Talmud of length 20 words or more, containing a total of 130,242 words (counting each pair of parallel passages a single time). This works out to an average of 3.3 parallel passages for each folio page of the Talmud.

The precision of our results is very high; inspection indicates that all parallels found by our method are genuine matches. In order to estimate recall, we compared the parallel passages identified by our algorithm with those found by a standard edit-distance algorithm which computes the Levenshtein distance for every pair of passages. To be sure, as noted above, it would be impractical to run an edit-distance algorithm across the entire Babylonian Talmud. Instead, we ran the edit-distance algorithm on a subset of the corpus, aiming to match passages in Tractate Bava Batra (89,000+ words) with passages in Tractate Gittin (61,000+ words), where the passages are both at least 20 words in length. Altogether, the Levenshtein-based algorithm identified 41 parallel passages between these two tractates (where passages were considered a match if their Levensthein distance was within 30% of the length of the first passage).

All 41 of these passages were correctly identified by our algorithm. Moreover, our algorithm identified an addition 5 valid parallels that were missed by the Levenshtein-based algorithm. These extra passages were cases in which interpolations of 4 or 5 words within one or the other of the passages caused the Levenshtein distance to rise above the threshold. In contrast, our algorithm is able to handle multi-word interpolations while still matching the skip-grams on either side. Thus, in addition to the gain in speed, our skip-gram-based algorithm also correctly identifies parallel passages which would be missed by a standard character-based edit-distance algorithm.

## III  AUTO-GENERATING A LEXICAL SUBSTITUTION LIST

As we have seen, our algorithm allows for word alternations and interpolations within the parallel passages. In such cases, we not only gain knowledge of the words that are similar to one another; rather, we also gain useful information with regard to the words that do not match each other. Specifically, in these cases, our algorithm serves to isolate pairs of disparate words that tend to be used in parallel with one another.

For instance, in a case where two parallel passages use two completely different words at one given position, our skip-grams will skip over the disparate words, matching the words beforehand and afterward. The discrepancy that remains – the length of a single word – contains a pair of words that potentially correlate with one another. A case in point would be the passage cited in the introduction to this paper, in which we found that the phrase שפסקו ממנה בעלי אמנה שנאמר in Tractate Hagigah is parallel to the phrase שפסקו ממנה אנשי אמנה שנאמר in Tractate Shabbat. These two phrases will line up via two skip-grams that each skip the third word. The match of these two skip-grams isolates the two omitted words, אנשי and בעלי, as a potential pair.

In such cases, there is a reasonable chance that the pair of words that comprise the discrepancy are words that are semantically similar, which should be treated as identical when searching for parallel passages. Although not all cases will qualify as such, it stands to reason that the recurrence of a given discrepancy in multiple passages within the corpus reflects the existence of an inherent connection between the two words. Thus, our algorithm tallies up a count for all one-word discrepancies, and then generates a lexical substitution list containing all of the word pairs whose frequency is above a predetermined threshold.

It is reasonable to assume that if these word equivalencies had been known to us during the initial stages of our analysis, we would have been able to widen our identification of parallel passages within the corpus. Furthermore, upon that basis, we would presumably be able to identify additional word equivalencies.

Therefore, upon generating the lexical substitution list, our algorithm reruns the entire procedure starting from step 1 above. This time, during step 1, while calculating the two-letter reductions for each word in the input corpus, the algorithm checks whether the word in question appears in the list, and if so, the algorithm stores a second alternate two-letter reduction for the given word, based upon the equivalent word in the list. Subsequently, when collecting the skip-grams for the corpus, the algorithm examines each group of 5 words to determine whether one or more of the words bears an alternate two-letter reduction (as per the list). If it does, the algorithm generates two separate sets of skip-grams for that group of words: one as usual, and one based upon the alternate two-letter reductions. Upon completing the full procedure, the resulting set of parallel passages is examined once again for one-word discrepancies. The lexical substitution list is updated accordingly, and the algorithm repeats again from step 1. This loop continues until we no longer see significant gains from one loop to the next.

Below are the results of running our algorithm on the Babylonian Talmud, after implementing the iterative lexical substitution list addition:

| Iteration | Number of entries in the list | Length of all Parallel Passages | Total Number of Parallel Passages |
|---|---|---|---|
| 1 | 0 | 130,242 | 4602 |
| 2 | 318 | 143,588 | 5272 |
| 3 | 396 | 143,631 | 5269 |

Table 5. Impact of Auto-generated Lexical Substitution List

As expected, the most significant gain was accrued during the second iteration, in which the total number of matched words rose over ten percent, and the total number of parallel passages rose nearly fifteen percent. The third iteration produced an additional minor gain of 43 words. The slight reduction in total passage count indicates that the extra words served to join parallel passages that would have otherwise been considered separate, effectively lowering the passage count while raising the total length. A fourth iteration provided no further gain.

## IV  CONCLUSIONS

In this paper we presented an effective method for identifying text reuse across large corpora of Hebrew and Aramaic texts – a task that was previously considered to be a formidable challenge fraught with difficulty.

Because our algorithm builds its list of potential matches from an index created via a single-pass preprocessing step, it can process texts of any size in O(N) time, allowing it to perform well on text with many millions of words. [3] This provides a significant performance increase over methods that rely on edit-distance calculation, in which the edit-distance needs to be computed separately for every pair of passages, resulting in $O(N^2)$ time. To be sure, the identification of clusters step in our algorithm is not linear, but since it is performed only on the set of potential matches, its impact upon the overall run time is relatively limited. Nevertheless, one can imagine bizarre cases in which this step could slow down the algorithm significantly. In future versions of the algorithm, we plan to improve the efficiency of this step, using heuristics to eliminate the need to test cases in which the probability of clustering is very low.

The use of n-gram matching followed by the identification of clusters of such matches is well attested in the literature [Grozea et al 2009]. The novelty here is in casting a wide net initially (matching only the two rarest letters and using skip-grams) and then tightening the requirements by seeking clusters of such matches. The use of 8-letter skip-grams simultaneously increases recall and allows efficient indexing.

The iterative generation of a lexical substitution list is of interest in its own right and improves results considerably. Of course, the identification of synonyms based on similarity of usage in corpora is a venerable idea [Harris, 1954]; for our purposes, we propose using substitutability in parallel texts rather than more general distributional similarity, as has been typical in previous work [Lee, 1999; Lowe, 2001].

The efficiency of the algorithm presented here means that when preparing Hebrew and Aramaic databases for use, queries regarding parallel passages do not have to be performed offline and stored for later use; rather, it is now practical to generate results dynamically on demand.

---

[3] To be sure, the identification of clusters step in our algorithm is not linear, but since it is performed only on the set of potential matches, it does not typically impact the overall run time significantly. Still, one can imagine bizarre cases in which this step could slow down the algorithm and would need to be executed in a more efficient manner.

                    http://jdmdh.episciences.org

Although the examples within this paper all focus upon the corpus of the Babylonian Talmud, the same method can be applied to other corpora of Hebrew and Aramaic texts as well, whether in an attempt to find all parallel passages throughout a given corpus, or in an attempt to find all parallel passages between two given corpora. Extending the algorithm to other semitic languages would require a small number of trivial changes in order to support a wider character range. For other languages, especially those of an analytic variety (such as English), with a relatively low morpheme-per-word ratio, the 4-out-of-5 skip-gram would likely be too broad; the frequent occurrences of articles and prepositions would render the skip-grams inefficient. For these languages, 5-out-of-6 or 6-out-of-7 skipgrams might be necessary.

The compiled executable is available for free download and reuse.[4]

---

[4] The command-line executable will be available and freely downloadable at [http1]. The executable is compiled for Windows, and assumes an installation of the Microsoft .NET framework. It takes a directory of text files as input and returns a text file detailing all of the parallel passages within the files. Running the executable with the -help parameter will display details of all required parameters.